\documentclass[sigconf]{acmart}

\AtBeginDocument{%
  \providecommand\BibTeX{{%
    \normalfont B\kern-0.5em{\scshape i\kern-0.25em b}\kern-0.8em\TeX}}}

\setcopyright{acmcopyright}
\copyrightyear{2021}
\acmYear{2021}
\acmDOI{10.1145/1122445.1122456}

\acmConference[MM'21]{Proceedings of the 29th ACM International Conference on Multimedia}{October 20--24, 2021}{Virtual Event, China.}
\acmBooktitle{Chengdu '21: Learning Human Motion Prediction via Stochastic Differential Equations,October 20--24, 2021, Chengdu, CN}
\acmPrice{15.00}
\acmISBN{978-1-4503-8651-7/21/10}

\begin{document}

\title{Learning Human Motion Prediction via Stochastic Differential Equations}
\thanks{Corresponding Authors: Zhenguang Liu, Haipeng Chen, Xuhong Zhang, Yuyu Yin.\\ Our code can be found at https://github.com/herolvkd/MM-2021-StochasticMotionPrediction}
\author{Kedi Lyu}
\email{lvkd19@mails.jlu.edu.cn}
\affiliation{%
	\institution{Jilin University}
	\streetaddress{Qianjin Street}
	\city{Changchun}
	\state{Jilin}
	\country{China}
	\postcode{130000}
}

\author{Zhenguang Liu}
\email{liuzhenguang2008@gmail.com}
\affiliation{%
	\institution{Zhejiang University}
	\city{Hangzhou}
	\state{Zhejiang}
	\country{China}
	\postcode{310000}
}
\author{Shuang Wu}
\email{wushuang@outlook.sg}
\affiliation{%
	\institution{Nanyang Technological University}
	\city{Singapore}
	\country{Singapore}
}
\author{Haipeng Chen}
\email{chenhp@jlu.edu.cn}
\affiliation{%
	\institution{Jilin University}
	\streetaddress{Qinajin Street}
	\city{Changchun}
	\state{Jilin}
	\country{China}
	\postcode{130000}
}
\author{Xuhong Zhang}
\email{zhangxuhong@zju.edu.cn}
\affiliation{%
	\institution{Zhejiang University}
	\city{Hangzhou}
	\state{Zhejiang}
	\country{China}
	\postcode{310000}
}
\author{Yuyu Yin}
\email{yinyuyu@hdu.edu.cn}
\affiliation{%
	\institution{Hangzhou Dianzi University}
	\city{Hangzhou}
	\state{Zhejiang}
	\country{China}
	\postcode{310000}
}

\begin{abstract}
 Human motion understanding and prediction is an integral aspect in our pursuit of machine intelligence and human-machine interaction systems. Current methods typically pursue a kinematics modeling approach, relying heavily upon prior anatomical knowledge and constraints. However, such an approach is hard to generalize to different skeletal model representations, and also tends to be inadequate in accounting for the dynamic range and complexity of motion, thus hindering predictive accuracy. In this work, we propose a novel approach in modeling the motion prediction problem  based on stochastic differential equations and path integrals. The motion profile of each skeletal joint is formulated as a basic stochastic variable and modeled with the Langevin equation. We develop a strategy of employing GANs to simulate path integrals that amounts to optimizing over possible future paths. We conduct experiments in two large benchmark datasets, Human 3.6M and CMU MoCap. It is highlighted that our approach achieves a $12.48\%$ accuracy improvement over current state-of-the-art methods in average.
\end{abstract}



\begin{teaserfigure}
  \centering
  \includegraphics[width=0.8\textwidth]{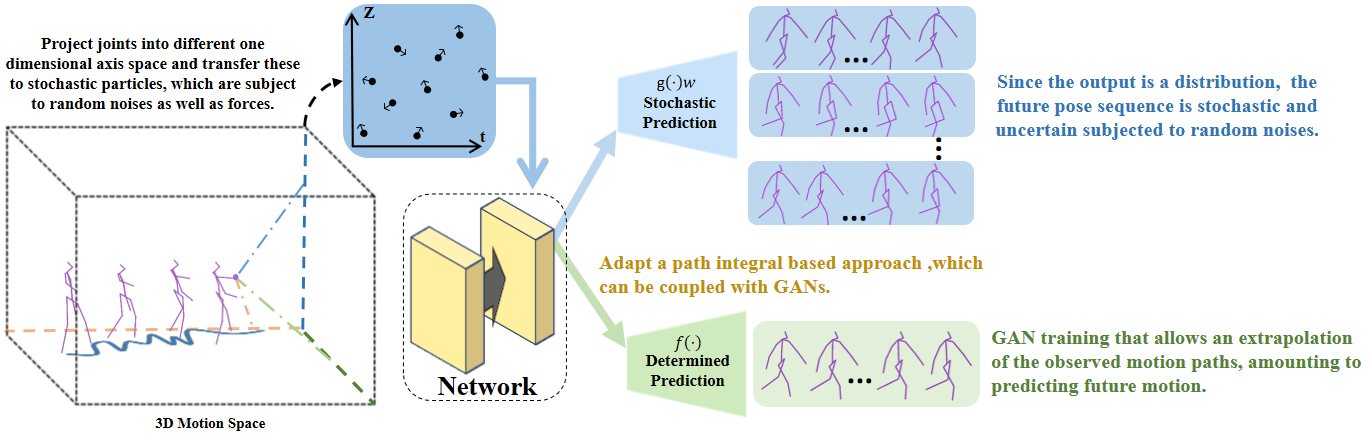}
  \caption{The process of modeling the human motion prediction based on stochastic differential equation and path integral. The network simulates path integrals that optimize possible paths of the stochastic particle and produce multiple predictions.}
  \label{fig:1}
\end{teaserfigure}

\maketitle

\section{Introduction}
An intuitive prediction of human actions in our surroundings greatly facilitates our daily interaction with the physical world. Developing machine models for human motion forecasting is likewise crucial in the fields of human-machine interactions and multimedia applications. Such models allow for a machine to properly respond to human actions and are also imperative for informed decision making and minimizing risks in safety-crucial applications \cite{,zangao18,shimingge} such as autonomous driving.

The key challenge of human motion prediction arises from the inherent stochastic nature of the problem. This is further compounded by the high dimensionality and complexity of human motion dynamics. Earlier works generally resorted to latent-variables models, such as hidden Markov models \cite{Markov} to adress the stochasticity of human motion. Fueled by the advances of deep learning, researchers adopted Recurrent Neural Networks (RNNs) such as Long Short-Term Memory (LSTM) \cite{erd}, Gated Recurrent Unit (GRU) \cite{residualgru} or refined hierarchical RNN architectures \cite{liu2019towards,Chen0YSZ20} that tackled the motion prediction task from a time series perspective. Kinematic trees and Graph Convolutional Networks (GCNs) have been employed to better model the structural correlations among skeletal joints \cite{trajectory,Alpher12,Alpher13,qiongjie,DGCN}.

Despite the promising results of recent deep learning based methods, their performance depends heavily on prior knowledge in the form of pre-defined skeletal structures or anatomical constraints. The rotational degrees of freedom of constituent skeletal joints have to be pre-specified. Otherwise, we observe a sharp decline in prediction accuracy. Furthermore, in the case of GCN based methods, the graphs have to be tuned accordingly and tailored to the kinematic structure of the skeleton to effectively learn the correlations between joints. Such heavy reliance on prior anatomical knowledge and necessity of engineering the network architecture accordingly to different skeletal structures constitute a severe limitation. This hinders the generalizability of current methods and also introduces systematic human bias when formulating the problem based on prior knowledge.

To address the above problem, as illustrated in Fig. \ref{fig:1}, we propose a reductionist approach where we strip the skeletal structure down to zoom in on the bare basic motion profiles of single skeletal joints. To effectively take into account the stochastic nature of the problem, we model each joint motion profile with a stochastic differential equation, namely the Langevin equation. The stochastic motion of each joint is subject to random noises as well as forces, constraints, and interactions with other joints that are learned from data. We then adopt a path integral based approach that can be coupled with Generative Adversarial Network (GAN) training.  This allows an extrapolation of the observed motion paths or trajectories, amounting to predicting future motion. Our approach has the additional benefit that the output is a distribution instead of a single deterministic output, which is aligned with the fact that the future is stochastic and uncertain.

To summarize, the main contributions of this paper are as follows: 1) We propose a novel formulation of the motion prediction task as a stochastic differential equation. 2) We develop a viable approach that employs GAN to simulate path integrals for solving the stochastic differential equations and predict future motion profiles. 3) Our method achieves new  state-of-art performance for both short-term and long-term motion predictions on two large-scale benchmark datasets: Human3.6M \cite{H36M} and CMU MoCap datasets \cite{CMU}. To the best of our knowledge, we are the first to combine stochastic differential equations with deep learning for human motion prediction. To facilitate future research, our code is released.

\section{Related Work}
{\bfseries Human motion prediction} \quad
Earlier works approached the human motion prediction task via hidden Markov models or Gaussian processes that encodes high dimensional skeletal motion data as parameterized latent variables. However, the hypothesis space of these latent variable models tend to be overly-restrictive and fails to capture the diverse range of motions faithfully, leading to inadequate performance. 
Driven by the advances of deep learning \cite{CuiZLYN20, FengHTC21, YangMM20, LiuSNK17, dong2021dual, liu2021toward, Gao2020Pairwise}, RNNs and sequence to sequence modeling inspired a major line of work \cite{erd,residualgru,liu2019towards,LiuL21} where various RNN architectures are utilized to better model motion dynamics.

Parallel to this development, researchers also proposed using kinematic trees \cite{liu2019towards}, Convolutional Neural Networks \cite{li2018convolutional,sunqr} and Graph Convolutional Networks \cite{trajectory,Alpher12,Alpher13,qiongjie} to model motion data, with the aim of better encapsulating the kinematics correlations amongst different skeletal joints. These efforts were effective in improving predictive accuracy. However, their implementations were generally dependent on imposing a skeletal structure and anatomical constraints which derives from prior knowledge. In particular, the rotational degrees of freedom of constituent joints have to be specified as constraints instead of learning such characteristics from data. Such imposed priors may also reflect human bias and prejudiced understanding, leading to systematic propagation of errors. In our work, we seek to mitigate this issue and learn joint correlations without imposing a prior structural constraint.

A concurrent track addresses the motion prediction problem with GANs \cite{,BarsoumKL18,KUN19,liu2021aggregated,zangao21,GAN-poster}. GANs carry the advantage of predicting a spectrum of possible futures instead of a single deterministic output, thus being more in line with the stochastic nature of the problem.

{\bfseries Stochastic Differential Equations} \quad
Many physical systems exhibit some extent of randomness. In part the stochastic behavior might be due to incomplete knowledge about the system. To a greater extent, this non-deterministic nature could be reflected by the inherent randomness in fundamental physics \cite{wolpert2008physical}, where even a hypothetical Laplace's demon with perfect knowledge would be unable to make deterministic predictions.

Modeling such stochastic physical systems began with the efforts to understand Brownian motion, \emph{i.e.}, the drift and diffusion of particles suspended in a liquid medium \cite{einstein1905motion}. Subsequent developments led to the Langevin equation \cite{langevin1908theorie}, a stochastic differential equation describing the evolution of such systems. The drift term in the Langevin equation corresponds to deterministic dynamical evolution in the system. The source of stochasticity lies in the diffusion term. It entails a white noise Gaussian process whose integral is known as the Wiener process \cite{wiener1957prediction}. The Langevin equation in its generic form is ambiguous and has to be complemented with a discretization scheme to have a well-defined notion of calculus \cite{van1992stochastic}, with the Ito discretization and Stranovich discretization constituting two major schemes.

Stochastic differential equations have been widely employed to model a plethora of physical phenomena \cite{van1992stochastic,moreno2015langevin} and systems with randomness such as quantitative finance and stock pricing \cite{hull1987pricing}. To the best of our knowledge, there are no previous works combining deep learning and stochastic differential equations for our motion prediction task. Motion prediction is of a stochastic nature and certainly falls within the paradigm of stochastic modeling. In the subsequent section, we outline our approach in leveraging the Langevin equation for modeling human motion.

\begin{figure*}
	\centering
	\includegraphics[width=0.8\textwidth]{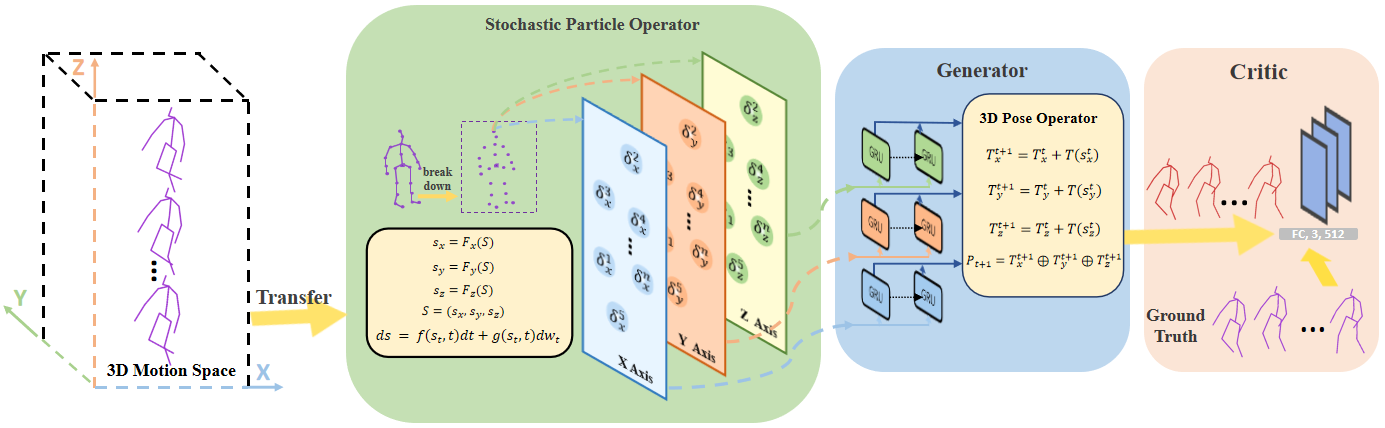}
	\caption{The outlined architecture of our approach.}
	\label{fig:2}
\end{figure*}

\section{Our Approach}

\subsection{Problem Formulation}
Given an observed 3D human pose sequence $\left(\mathbf{p}_{1},\mathbf{p}_{2},\cdots,\mathbf{p}_{K}\right)$, our intention is to extrapolate the future motion sequence $(\mathbf{p}_{K+1},\mathbf{p}_{K+2},$  $\cdots,\mathbf{p}_{K+T})$. Each pose $\mathbf{p}_t$ consists of skeletal joints $\mathbf{p}_t = \left[\mathbf{p}_t^1,\cdots, \mathbf{p}_t^N\right]$, where $N$ denotes the total number of joints and $\mathbf{p}_t^i\equiv(x_t^i,y_t^i,z_t^i)$ denotes the 3D coordinates of joint $i$ at time step $t$.

Contrary to previous works, we refrain from imposing prior constraints such as limiting rotational degrees of freedom in our problem formulation. Specifically, we strip away prior structures and consider the motion profiles of each skeletal joint $i$ as individual stochastic random variables $X^i,Y^i,Z^i$ respectively for the $x,y,z$ axis. Overall, the set of random variables $\mathbf{S}\equiv\{X^1,Y^1,Z^1,\cdots,X^N,Y^N,Z^N\}$ can be characterized as a system of Langevin equations as follows:
\begin{equation}
	\begin{aligned}  \label{eqn:sde}
		d\mathbf{S}^i(t)=f^i[\mathbf{S}(t)]dt+\sum_jg^{ij}[\mathbf{S}(t)]dW^j(t).
	\end{aligned}
\end{equation}
In Eqn. \ref{eqn:sde}, the term $f^i[\mathbf{S}(t)]$ describes the evolution of $\mathbf{S}^i$ whereby the overall dynamics of the entire system interaction is accounted for. Symbol $g^{ij}[\mathbf{S}(t)]$ denotes  the coupling of $\mathbf{S}^i$ to $W^j(t)$. For each superscript $j$, $W^j(t)$ represents an independent Wiener process \cite{wiener1957prediction} which constitutes a source of stochasticity in the system.

Intuitively, the $f^i[\mathbf{X}(t)]$ corresponds to a drift term that characterizes the dynamical evolution of the system deterministically whereas the $\sum_jg^{ij}[\mathbf{X}(t)]dW^j(t)$ corresponds to a diffusion term which exhibits stochastic behaviour. The coupling strength $g^{ij}$ determines the extent to which the joint trajectory is stochastic.

We now discretize Eqn. \ref{eqn:sde} for a time step $\Delta t $ in the Ito discretization scheme. This amounts to replacing $dW^j(t)$ with $w_t^j\sqrt{\Delta t}$ \cite{cugliandolo2017rules} where $w_t^j$ is a white noise Gaussian vector. In other words, $w_t^j$ has zero mean. The mean of $w_s^jw_t^j$ is one when $s=t$ and zero otherwise. Thus, we obtain:
\begin{equation}
	\begin{aligned} \label{eqn:discrete_sde}
		\mathbf{S}_t^i-\mathbf{S}_{t-1}^i=f_t^i\Delta t+\sum_jg_t^{ij}w_t^j\sqrt{\Delta t}.
	\end{aligned}
\end{equation}
In Eqn. \ref{eqn:discrete_sde}, the drift vector $f_t^i$ and the diffusion matrix $g_t^{ij}$ correspond to trainable network parameters.

\subsection{Path Integrals and Trajectory Probabilities}
For stochastic variable $\mathbf{S}^i$, the probability of observing a trajectory given an initial configuration $\mathbf{S}_I^i$ and final configuration $\mathbf{S}_F^i$ is formally given by a path integral \cite{adib2008stochastic}:
\begin{equation}
	\begin{aligned} \label{eqn:path_integral}
		\mathcal{P}(\mathbf{S}_F^i|\mathbf{S}_I^i)=\mathcal{J}\int_{\mathbf{S}_I^i}^{\mathbf{S}_F^i}e^{-\mathcal{S}[\mathbf{S}^i(t)]}\mathcal{D}[\mathbf{S}^i(t)].
	\end{aligned}
\end{equation}
Eqn. \ref{eqn:path_integral} integrates over all possible paths satisfying the given boundary conditions. $\mathcal{J}$ is a normalization factor. The term $\mathcal{S}[\mathbf{S}^i(t)]$ is the action functional (which is the analog of the action integral in Lagrangian mechanics) and is given by the integral of the Onsager–Machlup functional $L[f,g]$ (which is the analog of the Larangian) over time:
\begin{equation}
	\begin{aligned} \label{eqn:action}
		\mathcal{S}[\mathbf{S}^i(t)]=\int L[f^i(\mathbf{S}(t)),g^i(\mathbf{S}(t))] dt.
	\end{aligned}
\end{equation}

It is not feasible however to directly adapt the neural network to evaluate the path integral in Eqn. \ref{eqn:path_integral}. Therefore, we work with our discretized Langevin equation in Eqn. \ref{eqn:discrete_sde}. We impose the smallest possible time step in our task to be 1, \emph{i.e.} $\Delta t \to 1$. We are given an observed sequence $(\mathbf{S}_1, \mathbf{S}_2, \cdots, \mathbf{S}_K)$. The probability of the stochastic trajectory passing each of these points can be formulated as:
\begin{equation}
	\begin{aligned} \label{eqn:observed}
		\mathcal{P}(\mathbf{S}_1, \mathbf{S}_2, \cdots, \mathbf{S}_K)&=\prod_{t=2}^{K}\mathcal{P}(\mathbf{S}_{t+1}|\mathbf{S}_{t})\\
		&=\prod_{i}\prod_{t=2}^{K}\psi\left(\mathbf{S}_t^i-\mathbf{S}_{t-1}^i-f_t^i-\sum_jg_t^{ij}w_t\right).
	\end{aligned}
\end{equation}
Here, $\psi$ is the Dirac delta distribution. In other words, the probability of such a trajectory is a product of Dirac delta distributions.

From Eqn. \ref{eqn:observed}, we may learn network parameters $f,g$ to ensure that the argument of the Dirac delta function converges to zero. In other words, given an observed sequence, we learn the parameters of the associated systems of Langevin equations, including the drift vector and the diffusion matrix. Once these network parameters are in place, we may readily predict the future motion.

In practice, instead of strictly imposing the Dirac delta constraints in Eqn. \ref{eqn:observed}, we relax the problem by instead introducing a loss given as follows:
\begin{equation}
	\begin{aligned}
		L_{Observed} = \sum_i\sum_{t=2}^K\lVert \mathbf{S}_t^i-\mathbf{S}_{t-1}^i-f_t^i-\sum_jg_t^{ij}w_t \rVert^2.
	\end{aligned}
\end{equation}
This loss, evaluated over the observed sequence, effectively introduces a regularization term over the network weights. It serves to facilitate the fitting of the network weights to match the observed data, supported by an underlying stochastic differential equation model.

\subsection{Architecture Outline}
The objective of our architecture is to model human motion prediction based on stochastic differential equations and path integrals. We develop a strategy of employing a GAN structure, which simulates path integral and trains stochastic variables from the stochastic differential equation.  
As shown in Fig. \ref{fig:2}, the overall framework of our approach consists of three key components: Stochastic particle operator, Generator, and Critic. The \textit{stochastic particle operator} transfers the 3D motion sequence to stochastic particles. Then, the particles are fed into the \emph{generator} to predict future particles, which are translated to 3D future poses by a 3D pose operator. Finally, the \emph{critic} constrains the quality of the generated poses towards approaching the ground truth.  

\textbf{Stochastic particle operator}\quad
As mentioned earlier, we focus on using stochastic differential equations to deal with the motion prediction problem. Human pose is a complicated stochastic system. Previous works tried to constrain the randomness of the system. Differently, we make full use of this randomness by adopting stochastic differential equations. In order to remove the constraints, a human pose $p_t$ ($p_t\in \{p\}_1^K$, $p_t \in R^3$) is first broken down into joints $J=\{j_p^1, j_p^2,...,j_p^N\}$, where $N$ is the number of the joints. For each joint, we project it onto three axes, \emph{i.e.,} $x$, $y$, and $z$ axes. This yields three stochastic variables $s_x$, $s_y$, and $s_z$ ($s_x,s_y,s_z \in R$). The stochatic variables are one-dimensional scalars. 
Then, based on the Langevin equation (a kind of stochastic differential equation) we can calculate the differences of projected variables in consecutive frames according to Eqn. \ref{eqn:discrete_sde}. 
As such, we obtain the stochastic particles $\Delta = ({\delta}_1,{\delta}_2,...,{\delta}_n)$, where $\delta_i$ is a stochastic particle. The stochastic particles $\Delta$ constitute the input of the generator.

\textbf{Generator}\quad 
Current methods heavily rely on the capacities of the networks to generate future poses. However, the network has to deal with high dimensional pose data and high stochastic motion nature, which brings inherent difficulties in training. To address this issue, we design a generator that includes two components, a GRU network and a 3D pose operator. 

\textit{The GRU network component}\quad The objective of the GRU network is to produce the future stochastic particles, where each particle is just a one-dimensional scalar. This greatly reduces the difficulty of network training.
	 
\textit{3D pose operator}\quad More interestingly, to constrain the accuracy loss of the generated human poses, we design a 3D pose operator. In the 3D space, the movements of an object along the three axes are concurrent. We take the movement along $x$ axis as an example to show our operator. Firstly, we utilize the trajectory function $T(\cdot)$ to map stochastic variable $s_x^t$ to $s_x^{t+1}$, which is given by:
\begin{equation}\label{eqn:st}
	s_x^{t+1}=T(s_x^t)=s_x^t+{\delta_{t:t+1}}
\end{equation}
where $\delta_i$ is a future stochastic particle, generated by the GRU network. Then, the stochastic variables in consecutive frames form the trajectory in one-dimention space:
\begin{equation} \label{equ:T}
	T_{x}^{1+T:k+T}=(s_{x}^{1+T},s_{x}^{2+T},...,s_{x}^{k+T})
\end{equation} 	
Finally, we can concatenate different trajectories $T_x$, $T_y$, and $T_z$ into 3D pose sequence $p_{1+T:K+T}$:
\begin{equation} \label{eqn: pT}
		p_{1+T:k+T} = T_x^{1+T:k+T} \oplus T_y^{1+T:k+T} \oplus T_z^{1+T:k+T}
\end{equation}
where $ \oplus $ denotes concatenation, $k$ denotes the number of frames for the poses.  

On the other hand, the output of our method is a distribution since we adopt a GAN architecture. This allows generating multiple possible future pose sequences from the same input. To generate multiple plausible futures, a Gaussian distribution $w_i$ is mapped to the same space as the inputs. Let $\Gamma$ represent the final function of our network, the process of our architecture can be succinctly expressed as $p_{K+1:{K+T}}=\Gamma(p_{1:K},w_{1:K})$. Each drawn value of $w_i$ will sample different valid future pose sequences from the given input $p_i$.

\textbf{Critic}\quad   
A three-layers fully connected feedforward network is used as the critic.  First, it measures the similitude of the distributions between generated pose and the ground truth. Second, the critic assesses whether the stochastic generations are natural and smooth.

\subsection{Loss Function}\quad 
We introduce a reconstruction loss, an observation loss, a bone length loss, and a GAN loss to train our network.

The {\bfseries Reconstruction loss} constrains the predicted output to approach the ground truth: 
\begin{equation}
	L_{reconstruction}= \sum_{i=1}^T{||({\mathbf p_i-\mathbf p_i'})||^2}
\end{equation}
where $p$ denotes the pose sequence ground truth, while $p'$ denotes the generated pose sequence.

The {\bfseries Observation loss} as defined in Eqn. \ref{eqn:observed}, updates the network weights (which may be separately viewed as drift parameters $f^i_t$ and diffusion matrix parameters $g^{ij}_t$) according to the observed sequence. Thus, instead of relying on the reconstruction loss to provide feedback on the network parameters, we also effectively engage the prior observed data to construct a viable model of the system dynamics and stochasticity.

{\bfseries Bone length loss} enforces the bone length invariance across frames. It is defined as:
\begin{equation}
	L_{Bone}=(NT)^{-1} \sum_{f=t+1}^{t+T}\sum_{N=1}^{n-1}{||\mathbf{B}^f_{N}-\mathbf{B'}^f_{N}||_2}
\end{equation}
where ${B}^f_{N} $ and ${B'}^f_{N} $ are the estimated bone length and ground truth bone length. $N$ is the number of bones and $f$ is the number of frames.

{\bfseries GAN loss}  A critic network learns $D^G$ the authenticity of pose, incurring a Wasserstein GAN loss as:
\begin{equation}
	L_{critc}= \mathbf ED^G	[P_{t+1:t+T}]-\mathbf ED^G[{P}'_{t+1:t+T}]
\end{equation}
where $P$ denotes the real future pose sequence and $P'$ is a generated pose sequence.

\begin{table*}
	\caption{Performance evaluation (in MPJPE) on the H3.6m dataset. The best results are highlighted in bold.}

	\centering
	\resizebox{\textwidth}{!}
	{
		\renewcommand\tabcolsep{7.0pt}
		\label{tab:tab01}
		\begin{tabular}{ccccccccccccccccccccl}
			\hline
			
			&\multicolumn{5}{c}{Directions}  & \multicolumn{5}{c}{Greeting}    & \multicolumn{5}{c}{Phoning}         & \multicolumn{5}{c}{Posing}      \\ \hline
			Time (ms) & 80 & 160 & 320 & 400 & 1,000 & 80 & 160 & 320 & 400 & 1,000 & 80 & 160 & 320 & 400 & 1,000 & 80 & 160 & 320 & 400 & 1,000 \\ 
			
			LSTM3LR \cite{erd}
			&46.6 & 52.3& 87.0 & 101.6 &  135.1 
			&21.0 & 61.5 & 73.1& 78.2 & 177.6 
			&39.8 & 56.4 & 68.8 & 70.1 & 133.3 
			&46.1 & 72.0 & 130.1 & 156.7 & 176.5  \\ 
			
			RRNN \cite{residualgru}
			&36.4 &56.6 & 80.3& 98.1&126.3 
			&36.8 &73.3 & 138.2& 155.6& 189.5
			&24.3 &42.3 & 72.6& 82.3& 124.2
			&26.7 &52.4 & 129.5& 159.4& 181.7 \\ 
			
			HPGAN \cite{BarsoumKL18}
			& 80.9& 101.3& 148.6& 168.8 &  234.6 
			&81.5 & 118.8 & 178.4& 200.1 & 258.6
			&78.8 & 100.3 & 152.7 & 179.0 & 244.2
			&75.5 & 107.4 & 168.3 & 178.0 & 250.1  \\ 
			
			CEM  \cite{li2018convolutional}
			&22.0& 37.2& 59.6& 73.4&115.8
			&24.5& 46.2& 90.0& 103.1& 147.3
			&17.2&29.7 & 53.4& 61.3& 114.0
			&16.1 &35.6& 86.2& 105.6& 187.4 \\ 
			
			BiGAN \cite{KUN19}
			& 22.0 & 37.5 & 58.9& 72.0 & 114.7
			& 24.6 & 45.8 & 89.9 & 103.0 & 148.1
			& 17.0 & 29.7 & 54.1 & 62.1 & 112.0 
			& 16.8 & 35.0 & 86.4 &105.6 & 187.0\\
			
			MHR \cite{liu2019towards}
			& 23.3 & 25.0& 47.2 & 61.5 &  116.9
			&12.9 & 31.9 & 55.6& 82.5 & 123.2
			&12.5 & 21.3 & 39.3 & 58.6 & 112.8
			&13.6 & 23.5 & 62.5 & 114.1 & 143.6 \\ 
			
			FC-GCN \cite{,trajectory}
			& 12.6& 24.4 & 48.2 & 58.4 & 105.8  
			&14.5& 30.5& 74.2& 89.0& 140.9
			&10.4&14.3& 33.1&39.7& 105.1
			&9.4 &23.9& 66.2& 82.9& 175.0 \\ 
			
			LDR \cite{qiongjie}
			& 13.1 & 23.7 & 44.5& 50.9 & \textbf{78.3}
			& 9.6 & 27.9 & 66.3 & 78.8 & 129.7
			& 10.4 & 14.3  & 33.1 & 39.7 & 85.8
			& 8.7 & 21.1 & 58.3 &81.9 & 133.7\\	
			
			DMGNN \cite{Alpher13}
			& 12.3& 23.8& 46.2& 55.5&90.3 
			& 14.0& 29.8& 74.0& 89.1& 140.2
			& 10.2& 14.0& 32.8& 40.0& 104.1
			&9.2 & 23.5& 65.0& 82.8& 170.2 \\ 
			
			ours (Ours) 
			& \textbf{7.1 }& \textbf{17.8} & \textbf{42.5 }& \textbf{50.0} & 85.8
			& \textbf{8.5} &\textbf{ 22.3} & \textbf{50.1} & \textbf{68.2} & \textbf{71.3}
			& \textbf{9.2} & \textbf{11.8 }& \textbf{28.9} & \textbf{37.1} & \textbf{80.1 }
			& \textbf{8.5} & \textbf{13.8}& \textbf{29.9} &\textbf{38.8} & \textbf{80.6}\\ 
			\hline
			
			& \multicolumn{5}{c}{Walking} & \multicolumn{5}{c}{Eating}
			& \multicolumn{5}{c}{Smoking} & \multicolumn{5}{c}{Discussion}  \\ 
			\hline
			Time (ms) & 80 & 160 & 320 & 400 & 1,000 & 80 & 160 & 320 & 400 & 1,000 & 80 & 160 & 320 & 400 & 1,000 & 80 & 160 & 320 & 400 & 1,000 \\ 
			
			LSTM3LR \cite{erd}
			&27.5 & 42.8& 80.2 & 95.5& 129.0
			&21.3 & 38.8& 78.1 & 91.4& 121.1
			&26.5 &42.3 & 88.9& 99.2& 129.2
			&29.9 &49.4& 85.5& 105.8 & 135.1 \\ 
			
			RRNN \cite{residualgru}
			&20.5 & 39.8& 78.2 & 90.3& 120.1 
			&17.5 & 34.3& 71.1 & 87.5& 117.6
			&22.4 &39.9 & 80.2 & 92.5& 119.2
			&25.8 &43.4& 83.5& 95.8 & 129.1 \\ 
			
			HPGAN \cite{BarsoumKL18}
			&70.1& 89.6& 98.2 & 121.0 &  145.2
			&64.1 & 78.4 & 99.9& 113.7 & 136.2
			&67.2 & 88.6 & 100.1 & 123.9 & 140.4
			&71.4 & 91.3 & 105.2 & 129.7 & 150.4  \\ 
			
			CEM \cite{li2018convolutional}
			& 17.1& 31.2& 53.8& 61.5& 89.2
			&13.7& 25.9& 52.5& 63.3& 74.4
			&11.1&21.0& 33.4&38.3& 52.2
			&18.9 &39.3& 67.7& 75.7& 123.9 \\ 
			
			BiGAN \cite{KUN19}
			& 17.5 & 31.3 & 53.9& 61.4 & 89.5 
			& 13.6 & 26.1 & 51.4 & 63.1 & 74.1
			& 11.0 & 21.0 & 33.1 & 38.2 & 50.1
			& 19.2& 39.0 & 67.7 &75.3 & 122.5\\
			
			MHR \cite{liu2019towards}
			& 17.2 & 31.4 & 53.5& 61.1 & 89.0 
			& 13.2 & 26.0 & 51.1 & 62.6 & 74.0
			& 10.3 & 20.5 & 33.0 & 37.2 & 49.1
			& 19.0& 38.8  & 67.3 & 75.0 & 121.5\\
			
			FC-GCN \cite{trajectory}
			& 8.9& 15.7& 29.2& 33.4&50.9 
			&8.8 & 18.9& 39.4& 47.2& 57.1
			&\textbf{7.8}&14.9& 25.3&28.7& 44.3
			&9.8 &22.1& 39.6& 39.9& 69.5\\ 
			
			LDR \cite{qiongjie}
			& 8.9 & 14.9 & 25.4 & 29.9 & 45.8
			& 7.6 & 15.9 & 37.2 & 41.7& 53.8 
			& 8.1  & 13.4 & 24.8 & \textbf{24.9} & 43.1
			& 9.4 & 20.3 & 35.2 &41.2 & 67.4\\	
			
			DMGNN \cite{Alpher13}
			& 8.9& 14.9& 29.0& 33.1&50.2 
			&8.7& 18.7& 39.5& 47.1 & 57.0
			&8.2& 14.5& 25.1& 58.8 & 56.8
			&9.7 &21.9& 39.5& 40.0 & 68.3 \\ 
			
			ours (Ours) 
			& \textbf{8.2} & \textbf{13.8} & \textbf{22.6} & \textbf{27.9} & \textbf{39.8}
			& \textbf{7.3} & \textbf{15.1}& \textbf{36.2} & \textbf{41.1} &\textbf{50.0}
			& \textbf{7.8} & \textbf{13.0}  & \textbf{24.0} & \textbf{24.9} & \textbf{41.5}
			& \textbf{9.1} & \textbf{19.2}& \textbf{31.1} &\textbf{39.3} & \textbf{59.8}\\ 
			\hline
			
			& \multicolumn{5}{c}{Purchases}&\multicolumn{5}{c}{Sitting} & \multicolumn{5}{c}{Sittingdown}&\multicolumn{5}{c}{Takingphoto} \\ \hline
			Time (ms) & 80 & 160 & 320 & 400 & 1,000 & 80 & 160 & 320 & 400 & 1,000 & 80 & 160 & 320 & 400 & 1,000 & 80 & 160 & 320 & 400 & 1,000 \\ 
			
			LSTM3LR \cite{erd}
			&39.1 & 78.6 & 88.0 & 104.1 & 142.9
			&34.0 & 57.1 & 115.0& 111.3 & 142.1 
			&36.9 & 63.0 & 88.1 & 121.3 & 198.6
			&35.4 & 47.5 & 71.1 & 74.0  & 155.6  \\ 
			
			RRNN \cite{residualgru}
			& 38.5& 70.1& 101.0& 102.3&131.2 
			&34.1 & 53.2& 110.4& 115.0& 150.1
			&28.6 &55.2& 85.6&115.8 & 180.0
			&23.1 &47.0& 92.3& 110.1& 149.2 \\ 
			
			HPGAN \cite{BarsoumKL18}
			& 42.4& 88.9& 95.0 & 120.2 &  170.2
			&36.3 & 60.0 & 120.0& 123.1 & 168.2
			&39.9& 65.9 & 92.1 & 130.0 & 200.2
			&38.0 & 49.3 & 79.9 & 83.8& 160.4  \\ 
			
			CEM \cite{li2018convolutional}
			& 29.4& 54.9& 82.2& 93.0&139.3
			& 19.8& 42.4& 77.0& 88.4& 120.7
			&17.1&34.9& 66.3&77.6& 150.3
			&14.0&27.2& 53.8& 66.2& 128.1 \\ 
			
			BiGAN \cite{KUN19}
			& 29.0 & 54.1 & 82.2 & 92.4& 139.0
			& 19.9 & 41.0 & 76.3 & 88.2 & 120.5 
			& 17.0 & 34.8  & 66.5& 76.9 & 152.0
			& 14.2 & 27.1 & 53.5 &66.1 & 128.0\\
			
			MHR \cite{liu2019towards}
			& 15.3 & 30.6& 64.7 & 73.9 &  122.7 
			&12.6 & 25.6 & 44.7& 60.7 & 118.4 
			&9.6 & 18.6 & 41.1 & 57.7 & 148.3
			&7.9 & 19.0 & 31.5 & 57.3 & 108.5\\ 
			
			FC-GCN \cite{trajectory}
			&19.6& 38.5& 64.4& 72.2&139.3
			&10.7& 24.6& 50.6& 62.0& 115.7
			&11.4&27.6& 56.4&67.6& 142.2
			&6.8 &15.2& 38.2& 49.6& 116.3\\ 
			
			LDR \cite{qiongjie}
			& 16.2 & 36.1 & 62.8 & 79.2 & 112.6
			& 9.2 & 23.1 & 47.2 & 57.7 & 106.5
			& 9.3  & 21.4  & 46.3& 59.3 & 144.6
			& 7.1 & 13.8 & 29.6 &44.2 & 116.4\\	
			
			DMGNN \cite{Alpher13}
			& 19.3& 38.0& 64.2& 72.1&112.3 
			&10.6 & 24.4& 50.3& 61.8& 115.3
			&11.2 &27.5&56.1&67.7& 143.3
			&7.1 &15.0& 38.1& 49.5& 116.4 \\ 
			
			ours (Ours) 
			&\textbf{ 11.9}& \textbf{27.8 }& \textbf{51.2} & \textbf{62.7} & \textbf{102.2}
			& \textbf{8.3} & \textbf{20.1} &\textbf{ 40.2} & \textbf{52.9} & \textbf{99.8} 
			& \textbf{8.9} & \textbf{17.8}  & \textbf{38.2} & \textbf{55.6} & \textbf{113.2}
			& \textbf{6.1}& \textbf{13.0} & \textbf{25.2} &\textbf{40.1} & \textbf{85.5}\\ 
			\hline
			
			& \multicolumn{5}{c}{Waiting}     & \multicolumn{5}{c}{Walkingdog} & \multicolumn{5}{c}{Walkingtogether} & \multicolumn{5}{c}{Average}     \\ \hline
			Time (ms) & 80 & 160 & 320 & 400 & 1,000 & 80 & 160 & 320 & 400 & 1,000 & 80 & 160 & 320 & 400 & 1,000 & 80 & 160 & 320 & 400 & 1,000 \\ 
			
			LSTM3LR \cite{erd}
			& 41.1 & 57.1& 100.1 & 120.4 &  159.0 
			& 77.1 & 80.1 & 160.1& 190.0 & 230.3
			&40.1 & 55.2 & 80.3 & 99.9 & 180.2 
			&37.5 & 56.9 & 88.0 & 202.4 & 156.3  \\ 
			
			RRNN \cite{residualgru}
			& 29.5& 60.4& 118.1& 138.5&165.3  
			&59.8& 78.6& 152.3& 178.3& 200.1
			&25.4& 53.2& 89.8&99.6& 183.4
			&30.0 &53.3& 98.9& 114.7& 151.1 \\ 
			
			HPGAN \cite{BarsoumKL18}
			& 65.2 & 98.1& 148.3 & 168.2 &  199.9
			&83.1 & 92.1& 170.0& 198.4& 238.8
			&68.6 & 79.9 & 95.3 & 108.4 & 188.1
			&64.2 & 87.3 & 123.5& 143.1 & 192.4 \\ 
			
			CEM  \cite{li2018convolutional}
			&17.9& 36.5& 74.9 & 90.7 &205.8
			&40.6& 74.7& 116.6& 138.7& 210.2
			&15.0& 29.9& 54.3 & 65.8 & 149.8
			&19.6 &37.8& 68.1& 80.2& 133.9\\ 
			
			BiGAN \cite{KUN19}
			& 17.8 & 36.4 & 74.4 & 90.2 & 188.9
			& 41.2 & 78.3 & 116.2 & 130.1 & 210.5
			& 14.8 & 30.1  & 54.2 & 65.1 & 150.2
			& 19.7 & 37.8 & 67.9 &79.3 & 132.5\\
			
			MHR \cite{liu2019towards}
			& 12.8 & 24.5& 45.2 & 85.1&  121.9
			&38.2 & 63.6 & 109.3& 125.6& 190.0
			&12.2 & 25.2 & 46.2 & 50.2 & 134.1
			&15.4 & 28.4 &52.8 & 132.9& 118.3  \\ 
			
			FC-GCN \cite{trajectory}
			& 9.5& 22.0& 57.5& 73.9&107.5 
			&32.2& 58.0& 102.2& 122.7& 185.4
			&8.9&18.4& 35.3&44.3& 102.4
			&12.1 &24.6& 50.7& 60.8& 110.5 \\ 
			
			LDR \cite{qiongjie}
			& 9.2 & 17.6 & 47.2 & 71.6 & 107.3 
			& 25.3 & 56.6 & 87.9 & 99.4 & 143.2
			& 8.2  & 18.1 & 31.2& 39.4 & 79.2
			& 10.7 & 22.5 & 45.1 &56.0 & 96.5\\	
			
			DMGNN \cite{Alpher13}
			& 9.6& 21.8& 56.9& 71.9&106.9
			&31.8& 58.3& 101.9& 122.4& 184.3
			&8.8&18.0& 35.5&44.2& 102.5
			&12.0 &24.3& 50.3& 62.4& 107.9 \\ 
			
			ours (Ours) 
			&\textbf{ 7.1} & \textbf{17.0} & \textbf{40.0} & \textbf{50.8} & \textbf{92.1} 
			& \textbf{18.1} & \textbf{38.2} & \textbf{69.3 }& \textbf{80.0} & \textbf{131.5}
			& \textbf{7.6}  & \textbf{17.1}  & \textbf{28.8} & \textbf{38.9} & \textbf{75.2}
			& \textbf{8.9} & \textbf{18.5}& \textbf{37.2} &\textbf{47.2} & \textbf{81.8}\\ 
			\hline
			
	\end{tabular}}
\end{table*}


\begin{table*}
	\caption{Comparisons of MPJPE between our method and the state-of-the-art methods on CMU MoCap dataset. The best results are highlighted in bold.}

	\centering
	\resizebox{\textwidth}{!}{
		\renewcommand\tabcolsep{5.0pt}
			\label{tab:tab02}
		\begin{tabular}{ccccccccccccccccccccc}
			\hline
			& \multicolumn{5}{c}{Basketball}  & \multicolumn{5}{c}{Basketball Signal}    & \multicolumn{5}{c}{Directing Traffic}         & \multicolumn{5}{c}{Jumping}      \\ \hline
			Time (ms) & 80 & 160 & 320 & 400 & 1,000 & 80 & 160 & 320 & 400 & 1,000 & 80 & 160 & 320 & 400 & 1,000 & 80 & 160 & 320 & 400 & 1,000 \\ 
			
			LSTM3LR \cite{erd}  &15.4& 37.1& 44.8& 49.5& 110.3 &33.7& 56.8& 65.3& 72.6 &89.5 &26.3& 48.6& 53.9& 80.0 &135.6 &52.5& 86.5& 134.8& 151.3& 199.8 \\ 
			
			RRNN \cite{residualgru} & 18.5& 33.9& 48.1& 49.0& 106.3 &12.9& 23.8& 40.2& 60.1& 77.5& 15.6& 30.1& 55.2 &66.1& 127.1& 36.1& 68.7& 125.0& 140.0& 192.6\\ 
			
			CEM \cite{li2018convolutional} & 16.5& 30.5& 47.2& 48.8& 91.5 &8.7& 16.3& 30.1& 37.8& 76.6& 10.6& 20.3& 38.7&49.0& 113.3& 22.4& 44.3& 87.3& 105.1& 156.3\\ 
			
			FC-GCN \cite{trajectory} &14.3& 25.5& 48.4& 62.6& 109.0 &3.5& 6.7& 12.0& 15.8 &54.4 &7.4& 15.5& 31.9& 42.5 &151.9 &16.9& 34.4& 76.3& 98.6& 164.4 \\ 
			
			LDR \cite{Alpher13} & 13.1& 22.1& 37.2& 55.8& 97.7 &3.4& 6.2& 11.2& 13.8& 47.3& 6.8& 16.3& 66.3 &78.8& 129.7& 11.0& 24.5& 65.7& 90.3& 189.1\\ 
			
			Ours & \textbf{10.5} & \textbf{19.3}& \textbf{35.5} & \textbf{46.8} & \textbf{89.3} 
			& \textbf{2.5} & \textbf{6.0 }& \textbf{10.9 }& \textbf{12.8} & \textbf{44.5}
			& \textbf{4.9}  & \textbf{9.8 } & \textbf{21.9} & \textbf{29.7} & \textbf{98.5}
			& \textbf{10.4}& \textbf{23.5} & \textbf{60.1} & \textbf{85.6} & \textbf{148.2}\\ 
			\hline
			
			& \multicolumn{5}{c}{Running}  & \multicolumn{5}{c}{Soccer}     & \multicolumn{5}{c}{Walking}    & \multicolumn{5}{c}{Washing Window} \\ \hline
			
			Time (ms) & 80 & 160 & 320 & 400 & 1,000 & 80 & 160 & 320 & 400 & 1,000 & 80 & 160 & 320 & 400 & 1,000 & 80 & 160 & 320 & 400 & 1,000 \\ 
			
			LSTM3LR \cite{erd}  &18.6& 20.2& 28.9& 38.2& 53.1 &
			19.6& 36.7& 87.2& 112.7 &129.9 &11.3& 15.5& 25.4& 35.1 &41.4 &8.1& 14.6& 35.2& 37.9 & 69.8 \\ 
			
			RRNN \cite{residualgru} & 17.4& 20.0& 27.3& 36.7& 50.2 &
			20.3 & 39.5& 71.3& 84.0& 129.6& 8.2& 13.7& 21.9 &24.5& 32.2 & 8.4& 15.8& 29.3& 35.4& 61.1\\ 
			
			CEM \cite{li2018convolutional} & 14.3& 16.3& 18.0& 20.2& 27.5 &12.1& 21.8& 41.9& 52.9& 94.6& 7.6& 12.5& 23.0&27.5& 49.8& 8.2& 15.9& 32.1& 39.9& 58.9\\ 
			
			FC-GCN \cite{trajectory}  &25.5& 36.7& 39.3& 39.9& 58.2 &11.3& 21.5& 44.2& 55.8 &117.5&7.7& 11.8& 19.4& 23.1 & 40.2 &5.9& 11.9& 30.3& 40.0& 79.3 \\ 
			
			LDR \cite{Alpher13} & 15.2& 19.7& 23.3& 35.8& 47.4 &10.3& 21.1& 42.7& 50.9& 91.4& 7.1& 10.4& 17.8 &20.7& 37.5& 5.8& 12.3& 27.8& 38.2& 56.6\\ 
			
			Ours & \textbf{11.1} & \textbf{14.3} & \textbf{17.2} & \textbf{18.8} & \textbf{25.4}
			& \textbf{7.9}& \textbf{15.1}& \textbf{30.5} & \textbf{41.2} & \textbf{85.3} 
			& \textbf{5.8}& \textbf{8.5} & \textbf{15.1} & \textbf{17.2} & \textbf{30.1} 
			& \textbf{4.5}& \textbf{9.2} & \textbf{26.1} & \textbf{32.2} & \textbf{55.1}\\ \hline
			
	\end{tabular}}
\end{table*}

\section{Expriments}
In this section, we evaluate our method over large benchmark datasets Human3.6M and CMU MoCap, and compare it with state-of-the-art approaches. 
\subsection{Datasets and experimental setup}
\textbf{Human3.6M (H3.6M)}\quad  
H3.6M dataset contains 3.6 million human poses as well as their corresponding images that are recorded by a vicon motion capture system. These poses are divided into 7 subjects and involve 15 different classes of actions, such as walking, greeting, and phoning. In each action, 32 skeletal joints are used to  characterize the human pose. Following the evaluation protocol of previous works \cite{liu2019towards,erd}, we removed duplicate points in the human pose, performed down sampling (to 25 frames per second) on the video. Subject 5 (S5) is utilized as the test set and the remaining 6 subjects (S1, S6, S7, S8, S9, S11) are used for training. 

\textbf{CMU motion capture (CMU MoCap)}\quad  
We also conducted experiments on the large CMU MoCap dataset, which is released by researchers from Carnegie Mellon University. The dataset is captured by 12 infrared cameras, which record the positions of 41 markers taped on the human body. Following previous works \cite{trajectory,liu2021aggregated}, we adopt the same training/test splits and 8 actions are selected as our samples, e.g., running, basketball, soccer, and jumping. 
\begin{figure*}
	\centering
	\includegraphics[width=0.75\textwidth]{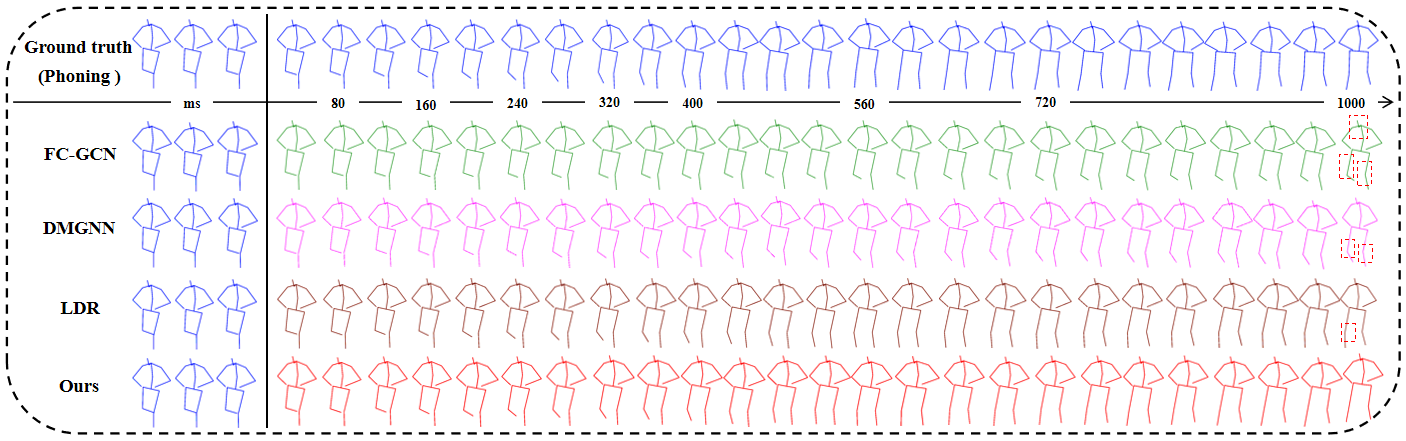}
	\caption{Visual comparisons on H3.6M dataset. The short-term predictions($<$$400ms$) and the long-term predictions($400$-$1,000ms$) are shown. The interval of animation is 40 $ms$.  The red box are unreasonable parts.}
	\label{fig:3}
\end{figure*}

\begin{figure*}
	\centering
	\includegraphics[width=0.75\textwidth]{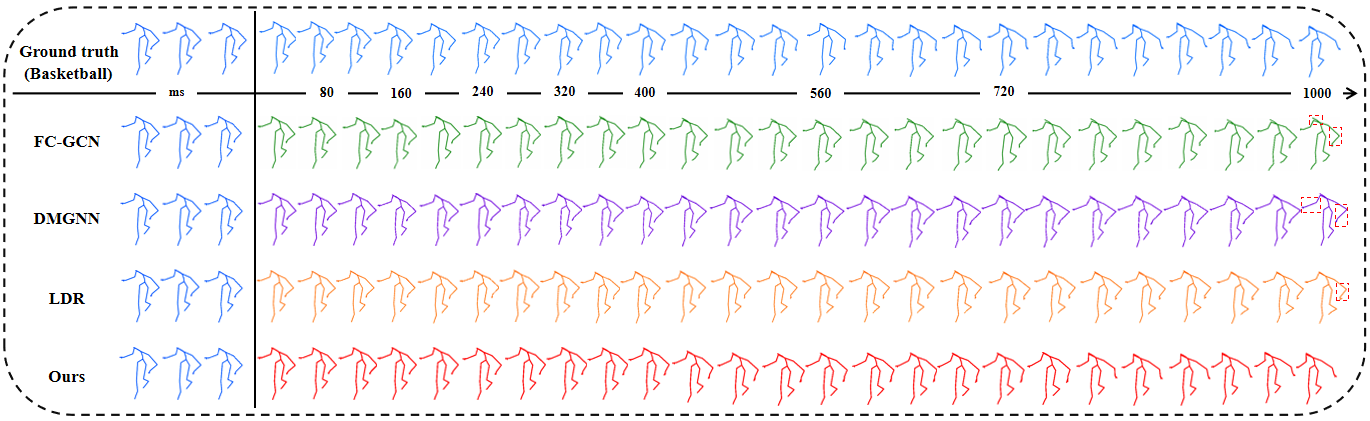}
	\caption{Visual comparisons on CMU MoCap dataset. The short-term predictions($<$$400ms$) and the long-term predictions($400$-$1,000ms$) are shown. The interval of animation is $40 ms$. The red box are unreasonable parts. }
	\label{fig:4}
\end{figure*}

\textbf{Experimental setup}
We use PyTorch to implement our method. The experiments were performed with a NVIDIA 1080Ti GPU. The hidden unit size of GRUs is 256. The Optimizer is Adam with a learning rate of 0.001, batch size is set to 16 for H3.6M and 32 for CMU MoCap. The critic is a three-layers fully connected feed-forward network with 256 units. The lengths of the historical sequence (input) and the predicted sequence (output) are set to 25 frames ($1,000ms$).

\subsection{Comparison with existing methods}
We compare our approach against existing state-of-the-art methods on the \textbf{H3.6m} and\textbf{ CMU MoCap} datasets. The performance is 
evaluated using the widely adopted metric MPJPE (Mean
Per Joint Position Error) in millimeter, \textit{i.e.}, the
spatial distance between ground truth and prediction.
The prediction results for 80, 160, 320, 400, 1,000 $ms$ are presented. By convention, we evaluate our approach on both short-term prediction (less than $400ms$) and long-term prediction ($400$-$1,000ms$). In addition to this, we also conducted experiments on \textbf{path search} and \textbf{stochastic prediction}.

{\bfseries Results on H3.6M}\quad
We benchmark our method against state-of-the-art methods, including LSTM3LR \cite{erd}, RRNN \cite{residualgru}, HP-GAN \cite{BarsoumKL18}, CEM \cite{li2018convolutional}, Bi-GAN \cite{KUN19}, HMR \cite{liu2019towards}, FC-GCN \cite{trajectory}, LDR \cite{qiongjie}, and DMGNN \cite{Alpher13}. These existing methods can be roughly classified into three categories: 1) RNN-based methods (LSTM3LR, RRNN, and CEM); 2) GAN-based methods (HP-GAN, Bi-GAN); and 3) pose representation-based methods (MHR, FC-GCN, LDR, and DMGNN). 

As shown in Table \ref{tab:tab01}, we compare the methods over all the 15 action types on the H3.6M dataset for both long-term and short-term predictions. These actions have different levels of complexity. 
For example, walking is a more regular action and demonstrates a high degree of periodicity whereas discussion is more complex and is generally aperiodic. 
From the quantitative results in Table \ref{tab:tab01}, we observe that our method achieves superior performance in quantitative accuracy over the state-of-the-arts for both long-term and short-term predictions. 

On the other side of the shield, by comparing with different categories of methods, we draw the following findings: 1) The poineering RNN-based methods perform relatively worse compared to other tested methods. This reveals the importance of effective pose representation schemas.  2) Pose representation-based methods such as LDR and DMGNN delivers better performance than other existing methods, while GAN based methods rank the second. 3) Our method significantly outperforms existing methods by a large margin, which confirms the efficacy of our method on the human motion prediction problem. 

Besides quantitative evaluation, we further compare the performance of state-of-the-art methods visually. As shown in Fig. \ref{fig:3}, we involve the results of the three state-of-the-art methods FC-GCN, MDGNN, and LDR, which achieve the best quantitative results among existing methods. In both short-term prediction and long-term prediction, our method is shown to be able to maintain a high consistency to the ground truth, and the human poses are more natural and smooth than existing methods.

{\bfseries Results on CMU MoCap}\quad  
Similarly, we also investigate our method on the CMU MoCap dataset with results reported in Table \ref{tab:tab02}. We compare our method with the state-of-the-art methods LSTM3LR, RRNN, CEM, FC-GCN, and LDR. In Table \ref{tab:tab02}, we show the comprehensive results on all the 8 actions. From the quantitative evaluation, we clearly observe that our method is able to deal with all kinds of actions in the CMU MoCap dataset and achieves much better performance consistently on all actions. The empirical results reconfirm the superiority of our approach on human motion prediction for both short-term and long-term predictions. Consistent and significant performance improvement over state-of-the-art methods on the two benchmark datasets evident that our method is robust. We further show the visual results on the CMU MoCap dataset in Fig. \ref{fig:4}. Similar to the observations on the H3.6M dataset, our method yields much enhanced visualization results. 

{\bfseries Path Search}\quad 
Our approach employs GANs to simulate path integrals that optimize over possible future paths. We show the visual process of simulating path integrals in Fig \ref{fig:5}. The left side of the figure shows the process of the path search. We demonstrate the results vertically according to the different paths. In each line, we show poses in different frames and present the prediction results for  80, 160, 320, 400, 560, 720, 1,000 (ms). The right side of the figure shows the comparison between the ground truth and the optimal path. From the results of the left and right sides, we can see that it is effective to simulate the path integral by adopting the GAN architecture.

\textbf{\begin{figure*}
		\centering
		\includegraphics[width=0.75\textwidth]{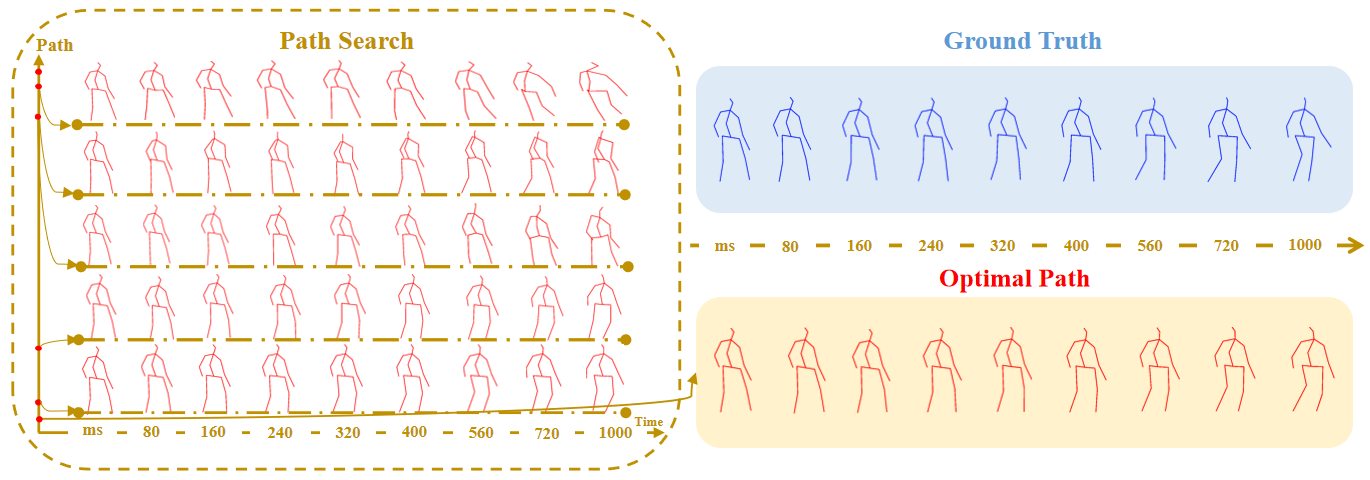}
		\caption{Visual illustration for path searching of Walktoghter. The results of different paths and the optimal paths for 80, 160, 320, 400, 560, 720, 1,000 $ms$ are presented.}
		\label{fig:5}
\end{figure*}}
\begin{figure*}
	\centering
	\includegraphics[width=0.75\textwidth]{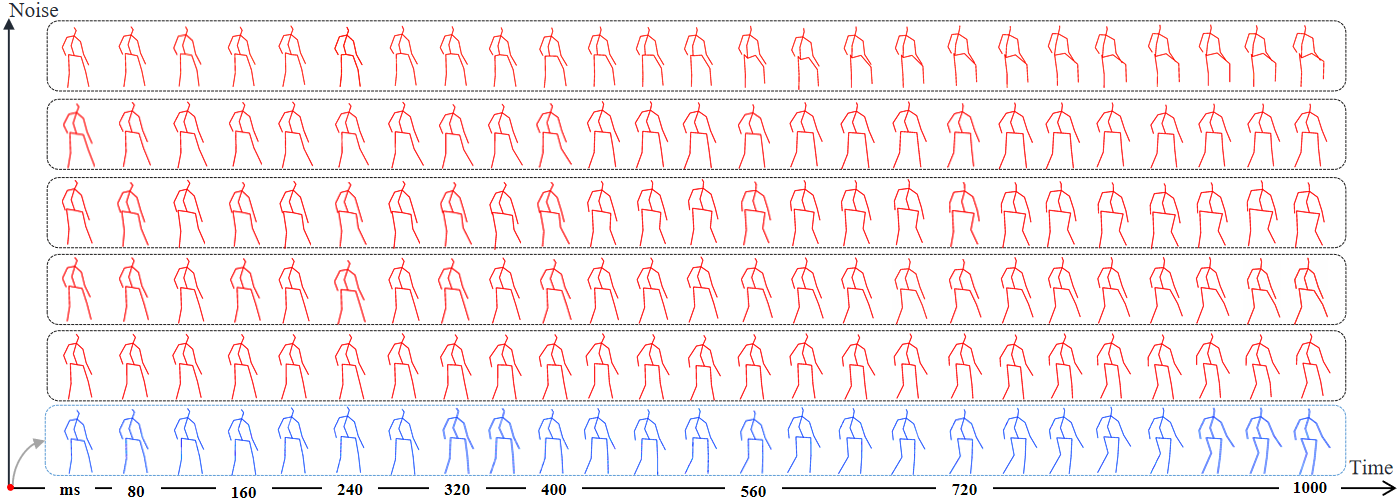}
	\caption{Visual illustration for stochastic predictions. The prediction results of Walktoghter with different noises are red and the ground truth is blue.}
	\label{fig:6}
\end{figure*}

{\bfseries Stochastic Prediction}\quad 
Interestingly, the output of our approach is a distribution rather than a determined pose sequence. Therefore, our approach is not only able to produce a determined result by searching the optimal path among different possibilities, but also could generate multiple stochastic futures. We generate stochastic predictions by adding noise to the input. As shown in Fig. \ref{fig:6}, we illustrate the results vertically according to the different noises. In each line, we show poses in different frames and present the prediction results for 80, 160, 320, 400, 560, 720, 1,000 ($ms$). We can see that our method is able to produce natural and smooth stochastic human motion sequences. 

\begin{table}[h]\footnotesize
	\caption{Ablation studies on different components of our
		approach. The method and its variants are evaluated over four actions from two datasets.}
	
	\setlength{\tabcolsep}{0.65mm}{
		\label{tab:tab03}
		\begin{tabular}{ccccccccccc}
			\hline
			& \multicolumn{5}{c}{Walking}  
			& \multicolumn{5}{c}{Disscussion}\\ \hline
			Time & 80 & 160 &320 & 400 & 1,000 & 80 & 160 & 320 & 400 & 1,000\\
			SPO
			& 69.8 & 88.5& 98.0 & 119.7 & 144.8
			& 71.0 & 90.9 & 105.0 & 129.2 & 149.9\\
			GRU
			&11.2& 17.3& 26.4& 32.5& 45.7
			&12.6& 24.3& 35.6& 43.7 &65.2\\
			3DPO
			& 19.4& 25.6& 37.1& 42.9& 59.2
			& 20.9& 33.4& 40.2& 48.8& 75.2\\
			Critic
			& 10.1& 16.9& 23.2& 30.1& 32.0
			& 10.9& 21.8& 23.6& 42.3& 63.1\\
			Ours
			& \textbf{8.2} & \textbf{13.8}& \textbf{22.6} &\textbf{ 27.9} & \textbf{39.8}
			& \textbf{9.1} & \textbf{19.2} & \textbf{31.1} & \textbf{39.3} & \textbf{59.8}\\
			\hline
			& \multicolumn{5}{c}{Basketball}         
			& \multicolumn{5}{c}{Jumping}\\ \hline
			SPO
			& 42.3 & 91.1& 123.1 &152.6& 210.2
			& 71.2 & 121.1& 198.2& 243.2& 326.8\\
			GRU
			&14.9& 25.6& 40.3& 51.2& 93.6
			&15.2& 25.8& 66.4& 91.2& 152.1 \\
			3DPO
			& 24.6& 35.7& 45.8 &59.8& 101.9
			& 23.7& 38.9& 77.9& 111.2& 162.4\\ 
			Critic
			& 12.6& 25.0& 43.2 &52.1& 99.3
			& 14.3& 25.9& 67.1& 99.9& 163.0\\ 
			Ours
			& \textbf{10.5}& \textbf{19.3}& \textbf{35.5} & \textbf{46.8} & \textbf{89.3}
			& \textbf{10.4} & \textbf{23.5}& \textbf{60.1 }& \textbf{85.6} & \textbf{148.2}\\ 
			\hline	
	\end{tabular}}
\end{table}

\subsection{Ablation Study}
In order to quantitatively analyze the contribution of different components in our method, we separately remove the Stochastic variable operator (\textbf{SPO}), \textbf{GRU}, 3D pose operator(\textbf{3DPO}), and the \textbf{Critic} from the method. 
The quantitative results are reported in Table \ref{tab:tab03}. (1) When we remove SPO, the method  performs like a classical GAN method  and the accuracy reduces greatly. 
(2) For the 3D pose operator, we utilize it to rebuild the human pose from stochastic particles. The results evident that the removal of 3D pose operator will lead to substantial accuracy reduction as well. Further, in order to study the effect of removing the GRU units, we replace it with an LSTM network. The results show that the performance drops slightly. This suggests that our choice of infrastructure is reasonable. 
(3) At last, we remove the critic and find that the prediction results are seriously affected, revealing the significance of the critic module.

\section{Conclusion}
In this paper, we have proposed a novel approach that model the human motion problem based on stochastic differential equations and path integrals. In our method, the motion profile of each skeletal joint is formulated as a basic stochastic variable and modeled with Langevin equation. GAN is utilized to simulate path integrals that optimize over possible paths. 
The experimental results show that our approach outperforms state-of-the-art methods by a large margin for both short-term prediction and long-term prediction. Overall, our work investigates the novel idea of utilizing stochastic differential equation and path integral to model human motion, and lays the theoretical foundations for this direction. We believe this is a solid step towards more accurate motion prediction and theoretical breakthroughs.

\section{Acknowledgement}
This research is partly supported by National Key Research and Development Program of China, Grant Number: 2018YFB0804202, 2018YFB0804203; Regional Joint Fund of NSFC, Grant Number: U19A2057; National Natural Science Foundation of China, Grant Number: 61876070; Jilin Province Science and Technology Development Plan Project, Grant Number: 20190303134SF.

\bibliographystyle{ACM-Reference-Format}
\bibliography{RE}
\end{document}